\documentclass[11pt,a4paper]{article}
\usepackage[hyperref]{naaclhlt2018}
\usepackage{times}
\usepackage{latexsym}

\usepackage{url}

\aclfinalcopy 

\usepackage[utf8]{inputenc}
\usepackage{times}
\usepackage{pifont}
\usepackage{xparse}

\usepackage{url}

\usepackage{color, colortbl}
\definecolor{SeaGreen}{rgb}{0.18, 0.55, 0.34}
\definecolor{Green}{rgb}{0.18, 0.55, 0.34}
\definecolor{Black}{rgb}{0.0, 0.0, 0.0}
\definecolor{Blue}{rgb}{0.0, 0.53, 0.74}
\definecolor{Orange}{rgb}{1.0, 0.4, 0.0}
\definecolor{Gray}{rgb}{0.66, 0.66, 0.66}
\definecolor{MidnightBlue}{rgb}{0.1, 0.1, 0.44}
\definecolor{White}{rgb}{1.0, 1.0, 1.0}

\usepackage{subcaption}
\usepackage{graphicx}
\usepackage{placeins}
\usepackage{float}

\usepackage{multirow}
\usepackage{booktabs}
\usepackage{tabularx}
\usepackage{collcell}
\usepackage{adjustbox}
\usepackage{array}
\usepackage{subcaption}
 
\newcommand*{\MinNumber}{0}%
\newcommand*{\MaxNumber}{1}%
 
\newcommand{\ApplyGradient}[1]{%
        \pgfmathsetmacro{\PercentColor}{100.0*(#1-\MinNumber)/(\MaxNumber-\MinNumber)}
        \hspace{-0.33em}\colorbox{MidnightBlue!\PercentColor!White}{}
}
 
\newcolumntype{R}{>{\collectcell\ApplyGradient}c<{\endcollectcell}}
\newcommand{\best}[1]{\cellcolor{Green!50}{\textbf{#1}}}
\newcommand{\sbest}[1]{\cellcolor{Green!26}{\textbf{#1}}}
\newcommand{\ssbest}[1]{\cellcolor{Green!7}{\textbf{#1}}}

\usepackage{latexsym}
\usepackage{amsmath}

\usepackage{enumitem}

\usepackage{xargs} 
\usepackage[colorinlistoftodos,prependcaption,textsize=tiny]{todonotes}
\newcommandx{\unsure}[2][1=]{\todo[linecolor=red,backgroundcolor=red!25,bordercolor=red,#1]{#2}}
\newcommandx{\change}[2][1=]{\todo[linecolor=blue,backgroundcolor=blue!25,bordercolor=blue,#1]{#2}}
\newcommandx{\info}[2][1=]{\todo[linecolor=OliveGreen,backgroundcolor=OliveGreen!25,bordercolor=OliveGreen,#1]{#2}}
\newcommandx{\improvement}[2][1=]{\todo[linecolor=Plum,backgroundcolor=Plum!25,bordercolor=Plum,#1]{#2}}
\newcommandx{\thiswillnotshow}[2][1=]{\todo[disable,#1]{#2}}

\usepackage{pgfplots}

\tikzstyle{block} = [rectangle, draw, fill=black!10, text width=7em,
  text centered, rounded corners, minimum height=4em] 
\tikzstyle{line}= [draw, -latex'] 
\usepackage{tikz} 
\usetikzlibrary{shapes,arrows}
\usepackage{forest} 
\forestset{ sn edges/.style={for tree={parent anchor=south, child
      anchor=north,align=center,base=bottom,where n
      children=0{tier=word}{}}, s sep=0mm} }

\title{Evaluating Discourse Phenomena \\in Neural Machine Translation}

\author{Rachel Bawden\textsuperscript{\normalfont 1}\qquad Rico Sennrich\textsuperscript{\normalfont 2,3}\qquad  Alexandra Birch\textsuperscript{\normalfont 2}\qquad  Barry Haddow\textsuperscript{\normalfont 2}\\
  \textsuperscript{1}LIMSI, CNRS, Univ. Paris-Sud, Université Paris-Saclay, F-91405 Orsay, France\\
  \textsuperscript{2}School of Informatics, University of Edinburgh, Scotland \\
  \textsuperscript{3}Institute of Computational Linguistics, University of Zurich, Switzerland \\
  {\tt rachel.bawden@limsi.fr} \\
  {\tt \{rico.sennrich, a.birch\}@ed.ac.uk} \\
 {\tt bhaddow@inf.ed.ac.uk}
}

\date{}

\begin{document}
\maketitle
\begin{abstract}
  For machine translation to tackle discourse phenomena, models must have access to extra-sentential
  linguistic context.  There has been recent interest in modelling context in neural machine
  translation (NMT), but models have been principally evaluated with standard automatic metrics,
  poorly adapted to evaluating discourse phenomena. In this article, we present hand-crafted,
  discourse test sets, designed to test the models' ability to exploit previous source and target
  sentences.  We investigate the performance of recently proposed multi-encoder NMT models trained
  on subtitles for English to French. We also explore a novel way of exploiting context from the previous
  sentence. Despite gains using \textsc{BLEU}, multi-encoder models give limited
  improvement in the handling of discourse phenomena: 50\% accuracy on our coreference test set
  and 53.5\% for coherence/cohesion (compared to a non-contextual baseline of 50\%). A simple
  strategy of decoding the concatenation of the previous and current sentence leads to good
  performance, and our novel strategy of multi-encoding and decoding of two
  sentences leads to the best performance (72.5\% for coreference and 57\% for coherence/cohesion),
  highlighting the importance of target-side context.
\end{abstract}

\section{Introduction}\label{sec:context}

Machine translation (MT) systems typically translate sentences independently of each
other. However, certain textual elements cannot be correctly translated
without linguistic context, which may appear outside the current sentence. The most obvious examples
of context-dependent phenomena problematic for MT are coreference
\citep{guillou_incorporating_2016}, lexical cohesion \cite{carpuat_one_2009} and lexical
disambiguation \cite{rios_gonzales_improving_2017}, an example for each of which is given in
(1-3). In each case, the English element in italic is ambiguous in terms of its French
translation. The correct translation choice (in bold) is determined by linguistic context
(underlined), which can be outside the current sentence. This disambiguating context can be source
or target-side; the correct translation of anaphoric pronouns \textit{it} and \textit{they} depends
on the gender of the translated antecedent (1). In lexical cohesion, a translation may depend on
target factors, but may also be triggered by source effects and linguistic mechanisms such as
repetition or alignment (2). In lexical disambiguation, source or target information may provide
the appropriate context (3).

\begin{enumerate}[label={(\arabic*)}, itemsep=0ex]
\small
\item The bee is busy. // \textit{It} is making honey.\mbox{}\\
 L'\underline{abeille}$_{\text{[f]}}$ est occupée. // \textbf{Elle}$_{\text{[f]}}$/\#il$_{\text{[m]}}$ fait du miel.
\item Do you fancy \underline{some soup?} // \textit{Some soup?} \mbox{}\\
Tu veux \underline{de la soupe}? // \textbf{De la soupe}/\#du potage?
\item And the \underline{code}? // Still some \textit{bugs}... \mbox{}\\
Et le \underline{code}~? // Encore quelques \textbf{bugs}/\#insectes...
\end{enumerate}

Recent work on multi-encoder neural machine translation (NMT) appears promising for the
integration of linguistic context
\cite{zoph_multi-source_2016,libovicky_attention_2017,jean_does_2017,wang_exploiting_2017}.  However
models have almost only been evaluated using standard automatic metrics, which are poorly adapted to
evaluating discourse phenomena. Targeted evaluation, in particular of coreference
in MT, has proved to be time-consuming and laborious \citep{guillou_incorporating_2016}.

In this article, we address the evaluation of discourse phenomena for MT and propose a novel
contextual model.  We present two hand-crafted, discourse test sets designed to test models'
capacity to exploit linguistic context for coreference and coherence/cohesion for 
English to French translation. Using these sets, we review contextual NMT strategies trained on subtitles in a
high-resource setting. Our new combination of strategies outperforms previous
methods according to our targeted evaluation and the standard metric \textsc{BLEU}.

\section{Evaluating contextual phenomena}\label{sec:evaluation}

Traditional automatic metrics are notoriously problematic for the evaluation of discourse in MT
\cite{hardmeier_discourse_2014}; discursive phenomena may have an impact on relatively few word
forms with respect to their importance, meaning that improvements are overlooked, and a correct
translation may depend on target-side coherence rather than similarity to a reference translation.

Coreference has been a major focus of discourse translation, spurred on by shared tasks on
cross-lingual pronoun prediction \cite{guillou_findings_2016,loaiciga_findings_2017}. Participants
were provided with lemmatised versions of reference translations,\footnote{This was to avoid
  pronoun forms being trivial to predict from the morphological inflections of other forms within
  the sentence, an unrealistic setting for MT output.} in which pronoun forms were to
be predicted. Evaluation in this setting (with the use of reference translations) was possible
with traditional metrics, because the antecedents were fixed in advance. However there are at least
two disadvantages to the approach: (i)~models must be trained on lemmatised data and
cannot be used in a real translation setting, and (ii)~many of the pronouns did
not need extra-sentential context; easier gains were seen for the
pronouns with intra-sentential antecedents and therefore the leaderboard was dominated by
sentence-level systems. 

\citeauthor{guillou_protest:_2016}'s (\citeyear{guillou_protest:_2016}) pronoun translation test suite
succeeds in overcoming some of these problems by creating an automatic evaluation method, with a
back-off manual evaluation.  Manual evaluation has always been an essential part of evaluating MT
quality, and targeted translation allows us to isolate a model's performance on specific
linguistic phenomena; recent work using in-depth, qualitative manual evaluation
\cite{isabelle_challenge_2017,scarton_quantitative_2015} is very
informative. \newcite{isabelle_challenge_2017} focus on specially constructed challenging examples
in order to analyse differences between systems.
They cover a wide range of linguistic phenomena, but since manual evaluation is costly and time-consuming,
only a few examples per phenomenon are analysed, and it is difficult to obtain quick, quantitative feedback.

An alternative method, which overcomes the problem of costly, one-off analysis, is to evaluate
models' capacity to correctly rank contrastive pairs of pre-existing translations, of which one is
correct and the other incorrect. This method was used by \newcite{sennrich_how_2017} to
assess the grammaticality of character-level NMT and again by
\newcite{rios_gonzales_improving_2017} in a large-scale setting for lexical disambiguation for
English-German.
The method allows automatic quantitative evaluation of specific phenomena at large scale, at the cost of only testing for very specific translation errors.
It is also the strategy that we will use here to evaluate translation of discourse phenomena.

\subsection{Our contrastive discursive test sets}

We created two contrastive test sets to help compare how well different contextual MT
models handle (i)~anaphoric pronoun translation and (ii)~coherence and
cohesion.\footnote{The test sets are freely available at \url{https://diamt.limsi.fr/eval.html}.} For each test set,
models are assessed on their ability to rank the correct translation of an ambiguous sentence higher
than the incorrect translation, using the disambiguating context provided in the previous source
and/or target sentence.\footnote{We acknowledge that in reality, the disambiguating context is not
  guaranteed to be in the previous sentence (cf. \citet[p.~161]{guillou_incorporating_2016}, for the
  distribution of intra- and inter-sentential anaphoric pronouns). However it is important to first
  judge in a controlled way whether or not models are actually capable of using extra-sentential
  linguistic context at all, before investigating longer distance context.} All examples in the 
test sets are hand-crafted but inspired by real examples from OpenSubtitles2016
\cite{lison_opensubtitles2016:_2016} to ensure that they are credible and that vocabulary and
syntactic structures are varied. The method can be used to evaluate any NMT model, by making it
produce a score for a given source sentence and reference translation.

Our test sets differ from previous ones in that examples necessarily need the previous context
(source and/or target-side) for the translations to be correctly ranked. Unlike the shared task test
sets, the ambiguous pronouns' antecedents are guaranteed not to appear within the current sentence,
meaning that, for MT systems to score highly, they must use discourse-level context. Compared to
other test sets suites, ours differs in that evaluation is performed completely automatically and
concentrates specifically on the model's ability to use context.  Each of the
test sets contains 200 contrastive pairs and is designed such that a non-contexual baseline system
would achieve 50\% accuracy.

\paragraph{Coreference test set}

\begin{figure}[ht]
\begin{center}
\includegraphics[width=\linewidth]{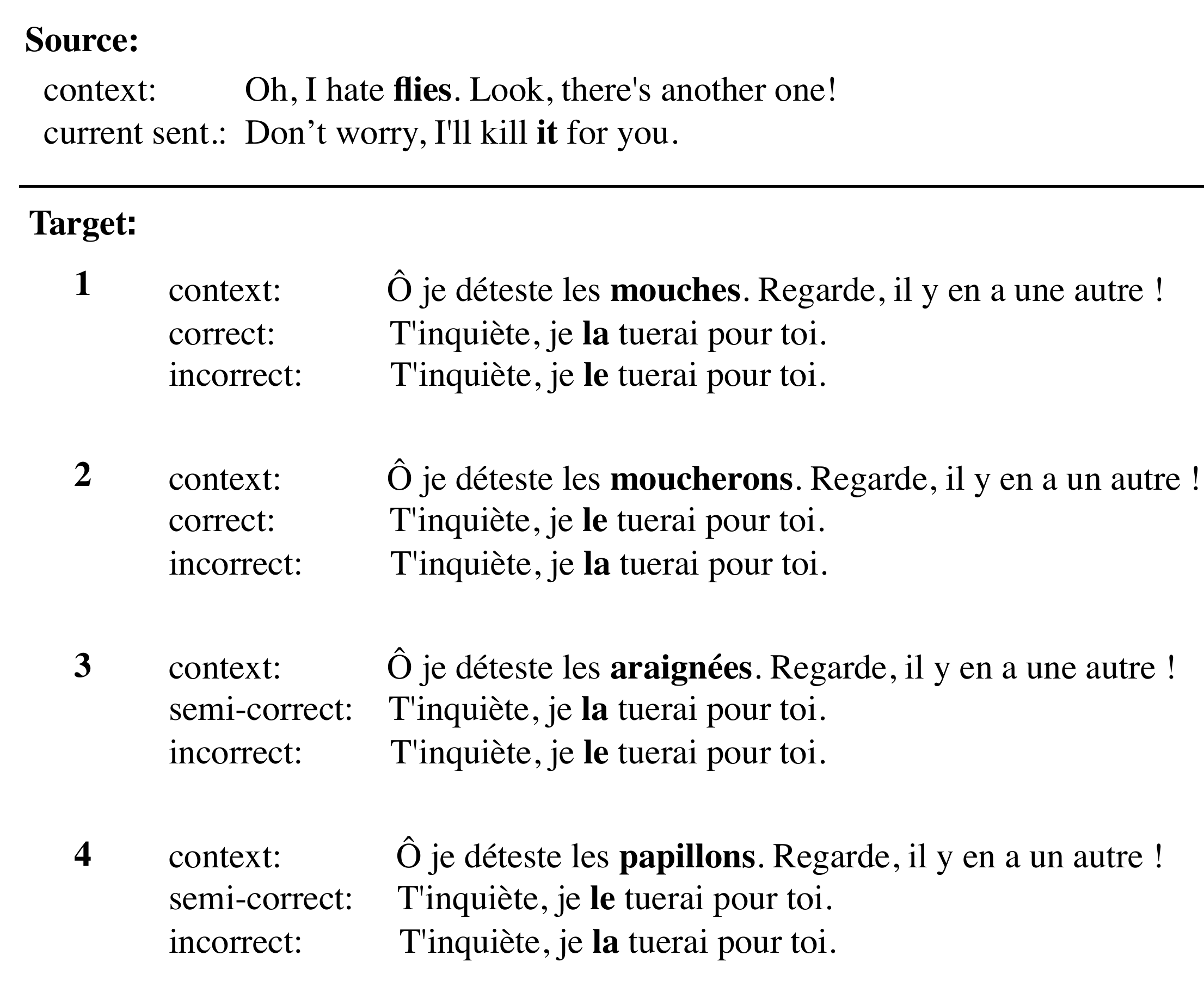}
\end{center}
\caption{Example block from the coreference set.}
\label{fig:testsetcoref}
\end{figure}

This set contains 50 example blocks, each containing four contrastive translation pairs (see the
four examples in Fig.~\ref{fig:testsetcoref}). The test set's aim is to test the integration of
target-side linguistic context. Each block is defined by a source sentence containing an occurrence
of the anaphoric pronoun \textit{it} or \textit{they} and its preceding context, containing the
pronoun's nominal antecedent.\footnote{The choice to use only nominal antecedents and only two
  anaphoric pronouns \textit{it} and \textit{they} is intentional in order to provide a controlled
  environment in which there are two contrasting alternatives for each example. This ensures that a
  non-contextual baseline necessarily gives a score of 50\%, and also enables us to explore this
  simpler case before expanding the study to explore more difficult anaphoric phenomena.} Four contrastive translation
pairs of the previous and current source sentence are given, each with a different translation of
the nominal antecedent, of which two are feminine and two are masculine per block. Each pair
contains a correct translation of the current sentence, in which the pronoun's gender is coherent
with the antecedent's translation, and a contrastive (incorrect) translation, in which the pronoun's
gender is inversed (along with agreement linked to the pronoun choice). Two of the pairs contain
what we refer to as a ``semi-correct'' translation of the current sentence instead of a ``correct''
one, for which the antecedent in the previous sentence is strangely or wrongly translated
(e.g. \textit{flies} translated as \textit{araignées} ``spiders'' and \textit{papillons}
``butterflies'' in Fig.~\ref{fig:testsetcoref}). In the ``semi-correct'' translation, the pronoun,
whose translation is wholly dependent on the translated antecedent, is coherent with this
translation choice. These semi-correct examples assess the use of target-side context, taking into
account previous translation choices.

Target pronouns are evenly distributed according to number and gender with 50 examples (25
correct and 25 semi-correct) for each of the pronoun types (m.sg, f.sg, m.pl and f.pl). Since there
are only two possible translations of the current sentence per example block, an MT
system can only score all examples within a block correctly if it correctly disambiguates, and a
non-contextual baseline system is guaranteed to score 50\%.

\paragraph{Coherence and cohesion test set}

\begin{figure}[ht]
\includegraphics[width=\linewidth]{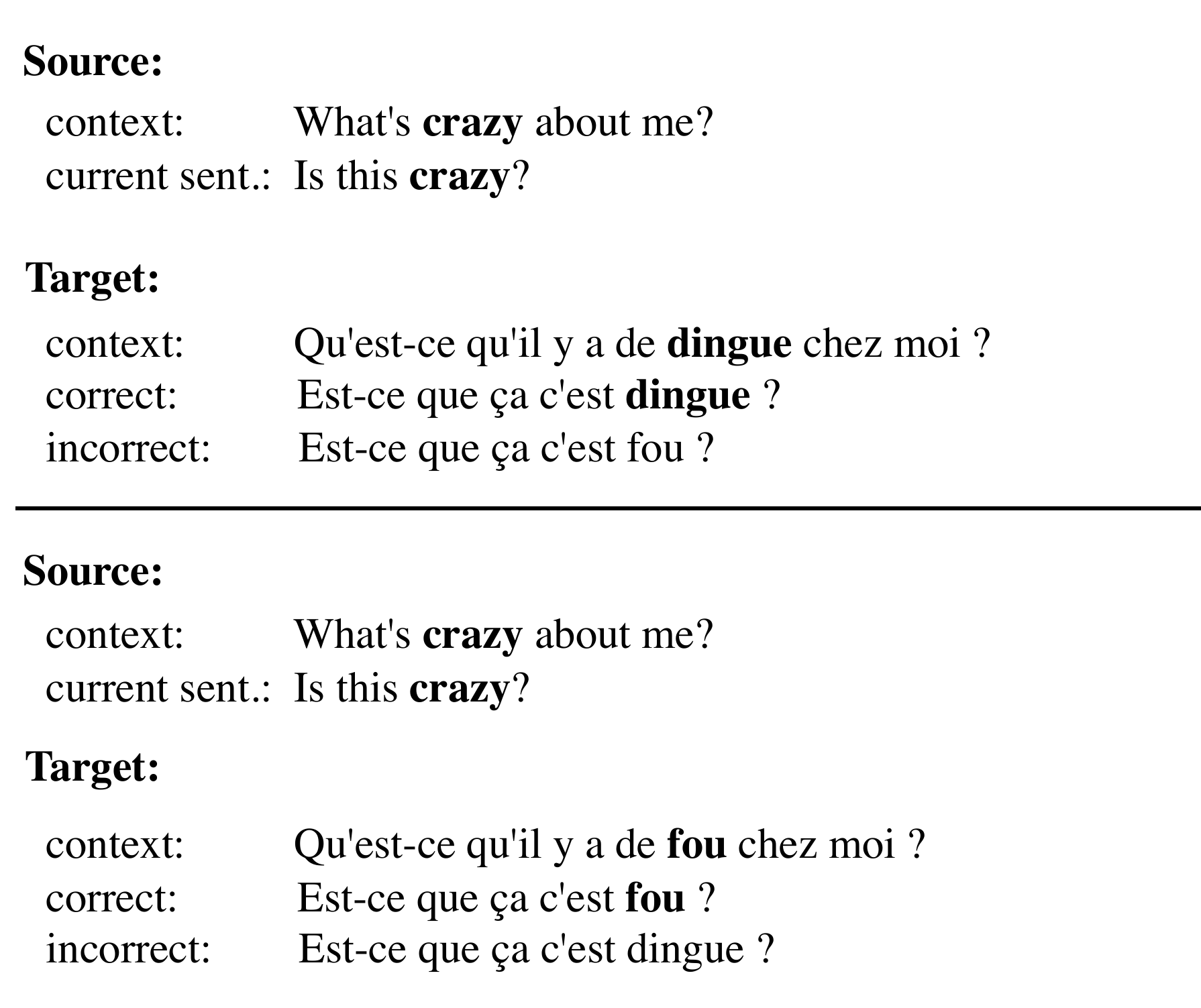}
\caption{Example block from the coherence/cohesion test: alignment.}
\label{fig:testsetlexical_repet}
\end{figure}

\begin{figure}[ht]
\includegraphics[width=\linewidth]{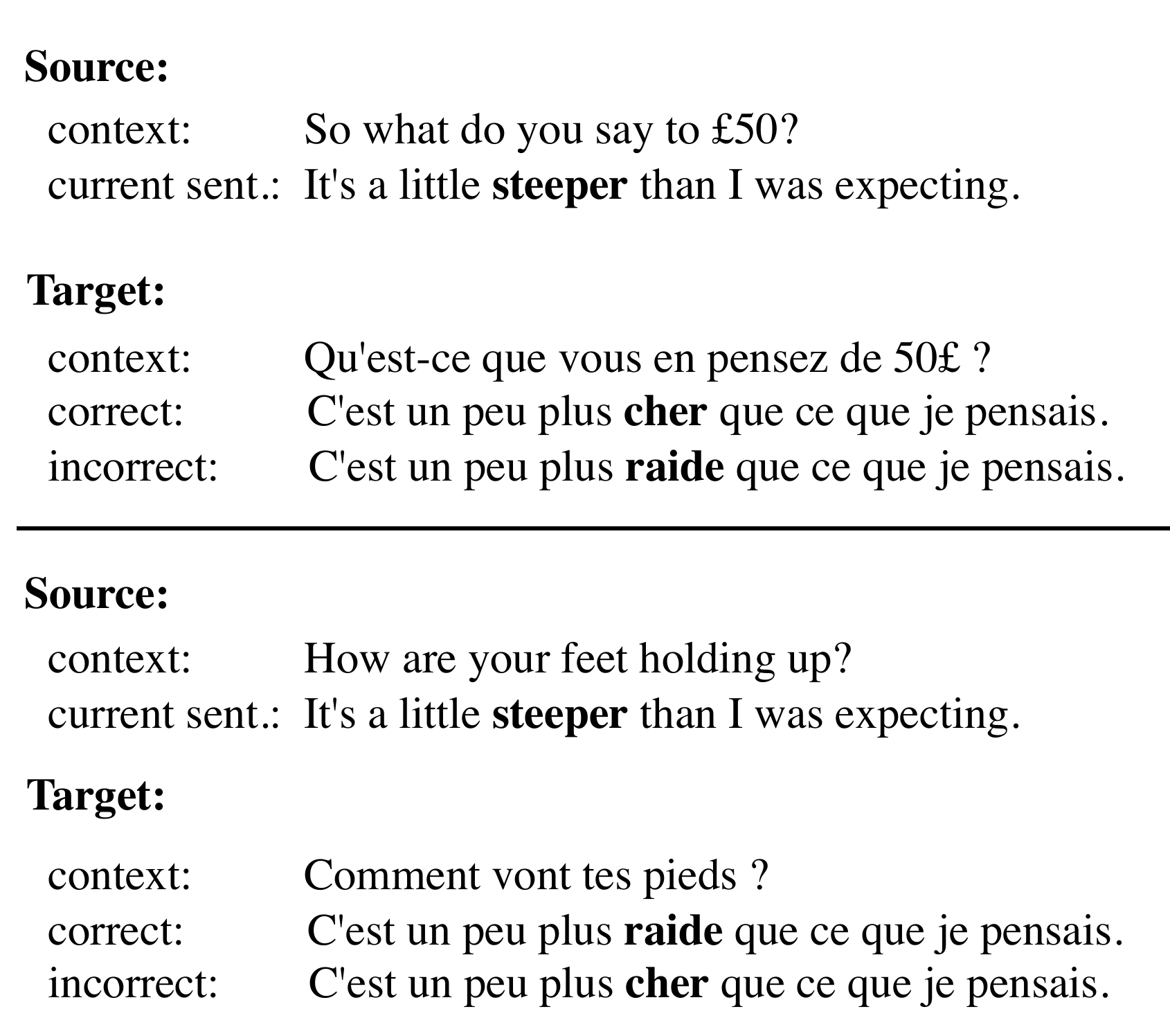}
\caption{Example block from the coherence/cohesion test: lexical disambiguation.}
\label{fig:testsetlexical_disambig}
\end{figure}

Coherence and cohesion concern the interpretation of a text in the context of discourse (i.e. beyond
sentence level). \newcite{de_beaugrande_introduction_1981} define the dichotomous pair as
representing two separate aspects: coherence relating to the consistency of the text to concepts and
world knowledge, and cohesion relating to the surface formulation of the text, as expressed through
linguistic mechanisms.

This set contains 100 example blocks, each containing two contrastive pairs (see
Figs.~\ref{fig:testsetlexical_repet} and \ref{fig:testsetlexical_disambig}). Each of the blocks is
constructed such that there is a single ambiguous source sentence, with two possible translations
provided. The use of one translation over the other is determined by disambiguation context found in
the previous sentence. The context may be found on the source
side, the target side or both. In each contrastive pair, the incorrect translation of the current
sentence corresponds to the correct translation of the other pair, such that the block can only be
entirely correct if the disambiguating context is correctly used.

All test set examples have in common that the current English sentence is ambiguous and that its
correct translation into French relies on context in the previous sentence. In some cases, the
correct translation is determined more by cohesion, for example the necessity to respect alignment
or repetition (Fig.~\ref{fig:testsetlexical_repet}). This means that despite two translations of an
English source word being synonyms (e.g.~\textit{dingue} and \textit{fou}, ``crazy''), they are not
interchangeable in a discourse context, given that the chosen formulation (alignment) requires
repetition of the word of the previous sentence. In other cases, lexical choice is
determined more by cohesion, for example by a general semantic context provided by the previous
sentence, in a more classic disambiguation setting as in Fig.~\ref{fig:testsetlexical_disambig},
where the English \textit{steeper} is ambiguous between French \textit{cher} ``more expensive'' and
\textit{raide} ``sharply sloped''. However, these types are not mutually exclusive and the
distinction is not always so clear.

\section{Contextual NMT Models}

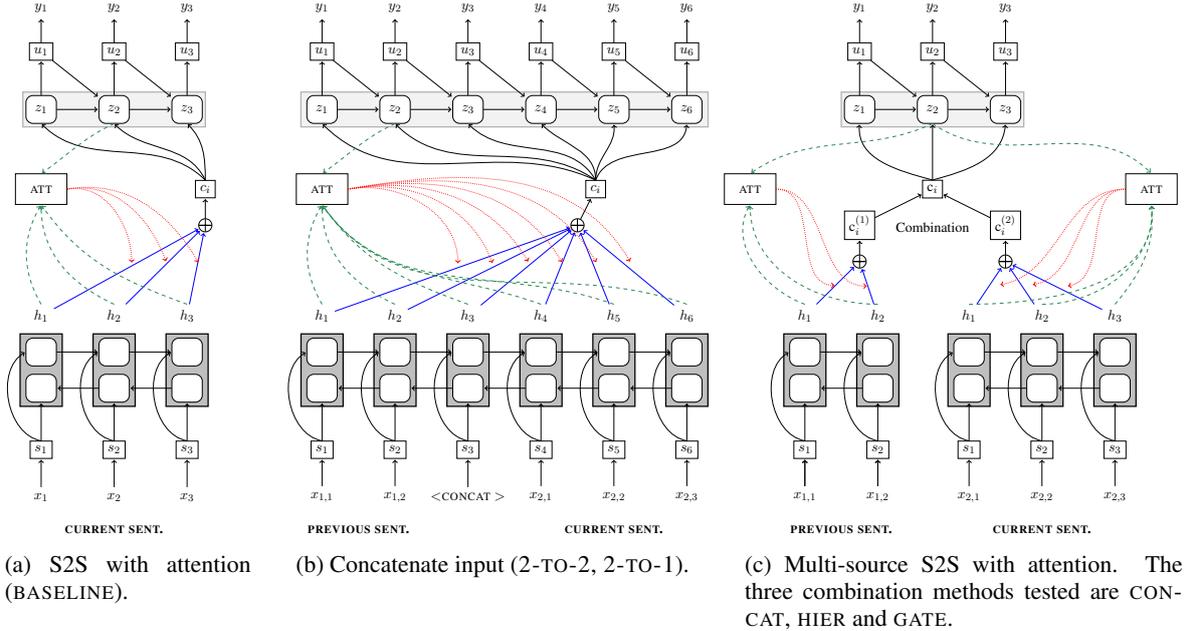
\begin{figure*}
 \captionsetup[subfigure]{width=0.9\textwidth}
\centering
\begin{subfigure}[t]{.225\textwidth}
\scalebox{.48}{
\begin{tikzpicture}[shorten >=1pt, auto, node distance=3cm, ultra thick,
   node_style/.style={circle,draw=blue,fill=blue!20!,font=\sffamily\Large\bfseries},
   edge_style/.style={draw=black, semithick}]
    \tikzstyle{every pin edge}=[<-,shorten <=1pt]
    \tikzstyle{neuron}=[circle,fill=black!25,minimum size=17pt,inner sep=0pt]
    \tikzstyle{input neuron}=[neuron, fill=green!50];
    \tikzstyle{output neuron}=[neuron, fill=red!50];
    \tikzstyle{block}=[rectangle, draw=black, thin];
    \tikzstyle{input} = [text centered]
    \tikzstyle{attention} = [block, text width=3em, text centered, minimum height=2.2em]
    \tikzstyle{rnn} = [block, text width=1.5em, text centered, fill=white, minimum width=1.5em,
                                minimum height=2em, rounded corners=.5em]
    \tikzstyle{rnn_block} = [block, line width=1pt, inner sep=0.2em,minimum width=3em,
                                minimum height=5.2em, fill=gray!50]
    \tikzstyle{decoder_block} = [block, line width=1pt, inner sep=0.2em,minimum width=13em,
                                minimum height=2.5em, fill=gray!10, draw=black!30]
    \tikzstyle{cross}=[circle, draw=black, thin, path picture={\draw[black]
        (path picture bounding box.south) -- 
        (path picture bounding box.north) (path picture bounding box.east) -- 
        (path picture bounding box.west)
        (path picture bounding box.north) (path picture bounding box.west) --
        (path picture bounding box.east);}]
    \tikzstyle{n}=[edge_style]
    \tikzstyle{bl}=[edge_style, bend left]
    \tikzstyle{br}=[edge_style, bend right]
    \tikzstyle{attedge}=[edge_style, draw=blue]
    \tikzstyle{alpha}=[edge_style, draw=red, densely dotted]
    \tikzstyle{attinput}=[edge_style, draw=SeaGreen, thick, dashed]

    \node[input] (x1) at (-2,-2){$x_1$};
    \node[input] (x2) at (0,-2){$x_2$};
    \node[input] (x3) at (2,-2){$x_3$};
                                
    \node[block] (s1) at (-2,-0.7){$s_1$};
    \node[block] (s2) at (0,-0.7){$s_2$};
    \node[block] (s3) at (2,-0.7){$s_3$};

    \draw[->, n] (x1) -- (s1){};
    \draw[->, n] (x2) -- (s2){};
    \draw[->, n] (x3) -- (s3){};

    \node[input] (xinput1) at (0,-2.9){\textbf{\textsc{current sent.}}};

    \node[rnn_block] (biblock1) at (-2,1.5){};
    \node[rnn] (biF1) at (-2,2) {};
    \node[rnn] (biB1) at (-2,1) {};

    \node[rnn_block] (biblock2) at (0,1.5){};
    \node[rnn] (biF2) at (0,2) {};
    \node[rnn] (biB2) at (0,1) {};

    \node[rnn_block] (biblock3) at (2,1.5){};
    \node[rnn] (biF3) at (2,2) {};
    \node[rnn] (biB3) at (2,1) {};

    \draw[->, n] (biF1) -- (biF2){};
    \draw[->, n] (biF2) -- (biF3){};
    \draw[->, n] (biB3) -- (biB2){};
    \draw[->, n] (biB2) -- (biB1){};

    \draw[->, bl] (s1.north) to [out=70,in=110] (biF1.west){};
    \draw[->, bl] (s2.north) to [out=70,in=110] (biF2.west){};
    \draw[->, bl] (s3.north) to [out=70,in=110] (biF3.west){};
    \draw[->, n] (s1) -- (biB1){};
    \draw[->, n] (s2) -- (biB2){};
    \draw[->, n] (s3) -- (biB3){};

    \node[input] (h1) at (-2,3){$h_1$};
    \node[input] (h2) at (0,3){$h_2$};
    \node[input] (h3) at (2,3){$h_3$};

    \node[decoder_block] (decoder) at (0,8.7){};
    \node[rnn] (z1) at (-2,8.7) {$z_1$};
    \node[rnn] (z2) at (0,8.7) {$z_2$};
    \node[rnn] (z3) at (2,8.7) {$z_3$};
    \draw[->, n] (z1) -- (z2){};
    \draw[->, n] (z2) -- (z3){};

    \node[block] (u1) at (-2,10.3){$u_{1}$};
    \node[block] (u2) at (0,10.3){$u_{2}$};
    \node[block] (u3) at (2,10.3){$u_{3}$};

    \draw[->, n] (z1) -- (u1){};
    \draw[->, n] (z2) -- (u2){};
    \draw[->, n] (z3) -- (u3){};

    \node[input] (y1) at (-2,11.5){$y_1$};
    \node[input] (y2) at (0,11.5){$y_2$};
    \node[input] (y3) at (2,11.5){$y_3$};

    \draw[->, n] (u1) -- (z2){};
    \draw[->, n] (u2) -- (z3){};
    \draw[->, n] (u1) -- (y1){};
    \draw[->, n] (u2) -- (y2){};
    \draw[->, n] (u3) -- (y3){};

    \node[attention] (att) at (-2,6.5){\textsc{att}};
    \node[cross] (cross) at (2.5,5.5){};
    \node[block] (C) at (2.5,6.5){$c_i$};

    \path[->, attedge] (h1) -- node [midway,right] (att1) {} (cross){};
    \path[->, attedge] (h2) -- node [midway,right] (att2) {} (cross){};
    \path[->, attedge] (h3) -- node [midway,right] (att3) {} (cross){};

    \draw[->, alpha] (att.east) to [out=10,in=90] (att1) {};
    \draw[->, alpha] (att.east) to [out=10,in=110] (att2) {};
    \draw[->, alpha] (att.east) to [out=10,in=120] (att3) {};

    \draw[->, n] (cross) -- (C){};
    \draw[->, n] (C.north) to [out=135,in=-45] (z1.south){};
    \draw[->, n] (C.north) to [out=115,in=-45] (z2.south){};
    \draw[->, n] (C.north) to [out=90,in=-45] (z3.south){};
    \draw[->, attinput] (z2.south) to [out=-155,in=35] (att.north){};
    \draw[->, attinput] (h1.north) to [out=120,in=-120] (att.south){};
    \draw[->, attinput] (h2.north) to [out=-205,in=-90] (att.south){};
    \draw[->, attinput] (h3.north) to [out=-200,in=-55] (att.south){};
\end{tikzpicture}}
\caption{S2S with attention (\textsc{baseline}).}
\label{fig:baseline}

\end{subfigure}
\begin{subfigure}[t]{.363\textwidth}

\scalebox{.48}{
\begin{tikzpicture}[shorten >=1pt, auto, node distance=3cm, ultra thick,
   node_style/.style={circle,draw=blue,fill=blue!20!,font=\sffamily\Large\bfseries},
   edge_style/.style={draw=black, semithick}]
    \tikzstyle{every pin edge}=[<-,shorten <=1pt]
    \tikzstyle{neuron}=[circle,fill=black!25,minimum size=17pt,inner sep=0pt]
    \tikzstyle{input neuron}=[neuron, fill=green!50];
    \tikzstyle{output neuron}=[neuron, fill=red!50];
    \tikzstyle{block}=[rectangle, draw=black, thin];
    \tikzstyle{input} = [text centered]
    \tikzstyle{attention} = [block, text width=3em, text centered, minimum height=2.2em]
    \tikzstyle{rnn} = [block, text width=1.5em, text centered, fill=white, minimum width=1.5em,
                                minimum height=2em, rounded corners=.5em]
    \tikzstyle{rnn_block} = [block, line width=1pt, inner sep=0.2em,minimum width=3em,
                                minimum height=5.2em, fill=gray!50]
    \tikzstyle{decoder_block} = [block, line width=1pt, inner sep=0.2em,minimum width=29em,
                                minimum height=2.5em, fill=gray!10, draw=black!30]
    \tikzstyle{cross}=[circle, draw=black, thin, path picture={\draw[black]
        (path picture bounding box.south) -- 
        (path picture bounding box.north) (path picture bounding box.east) -- 
        (path picture bounding box.west)
        (path picture bounding box.north) (path picture bounding box.west) --
        (path picture bounding box.east);}]
    \tikzstyle{n}=[edge_style]
    \tikzstyle{bl}=[edge_style, bend left]
    \tikzstyle{br}=[edge_style, bend right]
    \tikzstyle{attedge}=[edge_style, draw=blue]
    \tikzstyle{alpha}=[edge_style, draw=red, densely dotted]
    \tikzstyle{attinput}=[edge_style, draw=SeaGreen, thick, dashed]

    \node[input] (x1) at (-2,-2){$x_{1,1}$};
    \node[input] (x2) at (0,-2){$x_{1,2}$};
    \node[input] (x3) at (2,-2){\textsc{\textless concat \textgreater}};
    \node[input] (x21) at (4,-2){$x_{2,1}$};
    \node[input] (x22) at (6,-2){$x_{2,2}$};
    \node[input] (x23) at (8,-2){$x_{2,3}$};
                                
    \node[input] (xinput1) at (-1,-2.9){\textbf{\textsc{previous sent.}}};
    \node[input] (xinput2) at (6,-2.9){\textbf{\textsc{current sent.}}};

    \node[block] (s1) at (-2,-0.7){$s_1$};
    \node[block] (s2) at (0,-0.7){$s_2$};
    \node[block] (s3) at (2,-0.7){$s_3$};
    \node[block] (s21) at (4,-0.7){$s_4$};
    \node[block] (s22) at (6,-0.7){$s_5$};
    \node[block] (s23) at (8,-0.7){$s_6$};

    \draw[->, n] (x1) -- (s1){};
    \draw[->, n] (x2) -- (s2){};
    \draw[->, n] (x3) -- (s3){};
    \draw[->, n] (x21) -- (s21){};
    \draw[->, n] (x22) -- (s22){};
    \draw[->, n] (x23) -- (s23){};

    \node[rnn_block] (biblock1) at (-2,1.5){};
    \node[rnn] (biF1) at (-2,2) {};
    \node[rnn] (biB1) at (-2,1) {};

    \node[rnn_block] (biblock2) at (0,1.5){};
    \node[rnn] (biF2) at (0,2) {};
    \node[rnn] (biB2) at (0,1) {};

    \node[rnn_block] (biblock3) at (2,1.5){};
    \node[rnn] (biF3) at (2,2) {};
    \node[rnn] (biB3) at (2,1) {};

    \node[rnn_block] (2biblock1) at (4,1.5){};
    \node[rnn] (2biF1) at (4,2) {};
    \node[rnn] (2biB1) at (4,1) {};

    \node[rnn_block] (2biblock2) at (6,1.5){};
    \node[rnn] (2biF2) at (6,2) {};
    \node[rnn] (2biB2) at (6,1) {};

    \node[rnn_block] (2biblock3) at (8,1.5){};
    \node[rnn] (2biF3) at (8,2) {};
    \node[rnn] (2biB3) at (8,1) {};

    \draw[->, n] (biF1) -- (biF2){};
    \draw[->, n] (biF2) -- (biF3){};
    \draw[->, n] (biB3) -- (biB2){};
    \draw[->, n] (biB2) -- (biB1){};
    \draw[->, n] (2biB1) -- (biB3){};
    \draw[->, n] (biF3) -- (2biF1){};
    \draw[->, n] (2biF1) -- (2biF2){};
    \draw[->, n] (2biF2) -- (2biF3){};
    \draw[->, n] (2biB2) -- (2biB1){};
    \draw[->, n] (2biB3) -- (2biB2){};

    \draw[->, bl] (s1.north) to [out=70,in=110] (biF1.west){};
    \draw[->, bl] (s2.north) to [out=70,in=110] (biF2.west){};
    \draw[->, bl] (s3.north) to [out=70,in=110] (biF3.west){};
    \draw[->, n] (s1) -- (biB1){};
    \draw[->, n] (s2) -- (biB2){};
    \draw[->, n] (s3) -- (biB3){};

    \draw[->, bl] (s21.north) to [out=70,in=110] (2biF1.west){};
    \draw[->, bl] (s22.north) to [out=70,in=110] (2biF2.west){};
    \draw[->, bl] (s23.north) to [out=70,in=110] (2biF3.west){};
    \draw[->, n] (s21) -- (2biB1){};
    \draw[->, n] (s22) -- (2biB2){};
    \draw[->, n] (s23) -- (2biB3){};

    \node[input] (h1) at (-2,3){$h_1$};
    \node[input] (h2) at (0,3){$h_2$};
    \node[input] (h3) at (2,3){$h_3$};
    \node[input] (h21) at (4,3){$h_4$};
    \node[input] (h22) at (6,3){$h_5$};
    \node[input] (h23) at (8,3){$h_6$};

    \node[decoder_block] (decoder) at (3,8.7){};
    \node[rnn] (z1) at (-2,8.7) {$z_1$};
    \node[rnn] (z2) at (0,8.7) {$z_2$};
    \node[rnn] (z3) at (2,8.7) {$z_3$};
    \node[rnn] (z21) at (4,8.7) {$z_4$};
    \node[rnn] (z22) at (6,8.7) {$z_5$};
    \node[rnn] (z23) at (8,8.7) {$z_6$};

    \draw[->, n] (z1) -- (z2){};
    \draw[->, n] (z2) -- (z3){};
    \draw[->, n] (z3) -- (z21){};
    \draw[->, n] (z21) -- (z22){};
    \draw[->, n] (z22) -- (z23){};

    \node[block] (u1) at (-2,10.3){$u_{1}$};
    \node[block] (u2) at (0,10.3){$u_{2}$};
    \node[block] (u3) at (2,10.3){$u_{3}$};
    \node[block] (u21) at (4,10.3){$u_{4}$};
    \node[block] (u22) at (6,10.3){$u_{5}$};
    \node[block] (u23) at (8,10.3){$u_{6}$};

    \draw[->, n] (z1) -- (u1){};
    \draw[->, n] (z2) -- (u2){};
    \draw[->, n] (z3) -- (u3){};
    \draw[->, n] (z21) -- (u21){};
    \draw[->, n] (z22) -- (u22){};
    \draw[->, n] (z23) -- (u23){};

    \node[input] (y1) at (-2,11.5){$y_1$};
    \node[input] (y2) at (0,11.5){$y_2$};
    \node[input] (y3) at (2,11.5){$y_3$};
    \node[input] (y21) at (4,11.5){$y_4$};
    \node[input] (y22) at (6,11.5){$y_5$};
    \node[input] (y23) at (8,11.5){$y_6$};

    \draw[->, n] (u1) -- (z2){};
    \draw[->, n] (u2) -- (z3){};
    \draw[->, n] (u1) -- (y1){};
    \draw[->, n] (u2) -- (y2){};
    \draw[->, n] (u3) -- (y3){};
    \draw[->, n] (u21) -- (y21){};
    \draw[->, n] (u22) -- (y22){};
    \draw[->, n] (u23) -- (y23){};
    \draw[->, n] (u3) -- (z21){};
    \draw[->, n] (u21) -- (z22){};
    \draw[->, n] (u22) -- (z23){};

    \node[attention] (att) at (-2,6.5){\textsc{att}};
    \node[cross] (cross) at (5,5.5){};
    \node[block] (C) at (5.5,6.5){$c_i$};

    \path[->, attedge] (h1) -- node [midway,right] (att1) {} (cross){};
    \path[->, attedge] (h2) -- node [midway,right] (att2) {} (cross){};
    \path[->, attedge] (h3) -- node [midway,right] (att3) {} (cross){};
    \path[->, attedge] (h21) -- node [midway,right] (att21) {} (cross){};
    \path[->, attedge] (h22) -- node [midway,right] (att22) {} (cross){};
    \path[->, attedge] (h23) -- node [midway,right] (att23) {} (cross){};

    \draw[->, alpha] (att.east) to [out=10,in=90] (att1) {};
    \draw[->, alpha] (att.east) to [out=10,in=110] (att2) {};
    \draw[->, alpha] (att.east) to [out=10,in=120] (att3) {};
    \draw[->, alpha] (att.east) to [out=10,in=120] (att21) {};
    \draw[->, alpha] (att.east) to [out=10,in=120] (att22) {};
    \draw[->, alpha] (att.east) to [out=10,in=120] (att23) {};

    \draw[->, n] (cross) -- (C){};
    \draw[->, n] (C.north) to [out=155,in=-45] (z1.south){};
    \draw[->, n] (C.north) to [out=145,in=-45] (z2.south){};
    \draw[->, n] (C.north) to [out=120,in=-45] (z3.south){};
    \draw[->, n] (C.north) to [out=90,in=-45] (z21.south){};
    \draw[->, n] (C.north) to [out=90,in=-85] (z22.south){};
    \draw[->, n] (C.north) to [out=70,in=-95] (z23.south){};
    \draw[->, attinput] (z2.south) to [out=-155,in=35] (att.north){};
    \draw[->, attinput] (h1.north) to [out=120,in=-120] (att.south){};
    \draw[->, attinput] (h2.north) to [out=-205,in=-90] (att.south){};
    \draw[->, attinput] (h3.north) to [out=-200,in=-55] (att.south){};
    \draw[->, attinput] (h21.north) to [out=-200,in=-55] (att.south){};
    \draw[->, attinput] (h22.north) to [out=-200,in=-55] (att.south){};
    \draw[->, attinput] (h23.north) to [out=-200,in=-55] (att.south){};

\end{tikzpicture}}

\caption{Concatenate input (\textsc{2-to-2}, \textsc{2-to-1}).}
\label{fig:concat}
\end{subfigure}
\begin{subfigure}[t]{.40\textwidth}

\scalebox{.48}{
\begin{tikzpicture}[shorten >=1pt, auto, node distance=3cm, ultra thick,
   node_style/.style={circle,draw=blue,fill=blue!20!,font=\sffamily\Large\bfseries},
   edge_style/.style={draw=black, semithick}]
    \tikzstyle{every pin edge}=[<-,shorten <=1pt]
    \tikzstyle{neuron}=[circle,fill=black!25,minimum size=17pt,inner sep=0pt]
    \tikzstyle{input neuron}=[neuron, fill=green!50];
    \tikzstyle{output neuron}=[neuron, fill=red!50];
    \tikzstyle{block}=[rectangle, draw=black, thin];
    \tikzstyle{input} = [text centered]
    \tikzstyle{attention} = [block, text width=3em, text centered, minimum height=2.2em]
    \tikzstyle{rnn} = [block, text width=1.5em, text centered, fill=white, minimum width=1.5em,
                                minimum height=2em, rounded corners=.5em]
    \tikzstyle{rnn_block} = [block, line width=1pt, inner sep=0.2em,minimum width=3em,
                                minimum height=5.2em, fill=gray!50]
    \tikzstyle{decoder_block} = [block, line width=1pt, inner sep=0.2em,minimum width=13em,
                                minimum height=2.5em, fill=gray!10, draw=black!30]
    \tikzstyle{cross}=[circle, draw=black, thin, path picture={\draw[black]
        (path picture bounding box.south) -- 
        (path picture bounding box.north) (path picture bounding box.east) -- 
        (path picture bounding box.west)
        (path picture bounding box.north) (path picture bounding box.west) --
        (path picture bounding box.east);}]
    \tikzstyle{n}=[edge_style]
    \tikzstyle{bl}=[edge_style, bend left]
    \tikzstyle{br}=[edge_style, bend right]
    \tikzstyle{attedge}=[edge_style, draw=blue]
    \tikzstyle{alpha}=[edge_style, draw=red, densely dotted]
    \tikzstyle{attinput}=[edge_style, draw=SeaGreen, thick, dashed]

    \node[input] (x1) at (-0.5,-2){$x_{1,1}$};
    \node[input] (x2) at (1.5,-2){$x_{1,2}$};
    \node[input] (x21) at (4,-2){$x_{2,1}$};
    \node[input] (x22) at (6,-2){$x_{2,2}$};
    \node[input] (x23) at (8,-2){$x_{2,3}$};
   
    \node[input] (xinput1) at (0.5,-2.9){\textbf{\textsc{previous sent.}}};
    \node[input] (xinput2) at (6,-2.9){\textbf{\textsc{current sent.}}};
                             
    \node[block] (s1) at (-0.5,-0.7){$s_1$};
    \node[block] (s2) at (1.5,-0.7){$s_2$};
    \node[block] (s21) at (4,-0.7){$s_1$};
    \node[block] (s22) at (6,-0.7){$s_2$};
    \node[block] (s23) at (8,-0.7){$s_3$};

    \draw[->, n] (x1) -- (s1){};
    \draw[->, n] (x2) -- (s2){};
    \draw[->, n] (x1) -- (s1){};
    \draw[->, n] (x2) -- (s2){};
    \draw[->, n] (x21) -- (s21){};
    \draw[->, n] (x22) -- (s22){};
    \draw[->, n] (x23) -- (s23){};

    \node[rnn_block] (biblock1) at (-0.5,1.5){};
    \node[rnn] (biF1) at (-0.5,2) {};
    \node[rnn] (biB1) at (-0.5,1) {};

    \node[rnn_block] (biblock2) at (1.5,1.5){};
    \node[rnn] (biF2) at (1.5,2) {};
    \node[rnn] (biB2) at (1.5,1) {};


    \node[rnn_block] (2biblock1) at (4,1.5){};
    \node[rnn] (2biF1) at (4,2) {};
    \node[rnn] (2biB1) at (4,1) {};

    \node[rnn_block] (2biblock2) at (6,1.5){};
    \node[rnn] (2biF2) at (6,2) {};
    \node[rnn] (2biB2) at (6,1) {};

    \node[rnn_block] (2biblock3) at (8,1.5){};
    \node[rnn] (2biF3) at (8,2) {};
    \node[rnn] (2biB3) at (8,1) {};

    \draw[->, n] (biF1) -- (biF2){};
    \draw[->, n] (biB2) -- (biB1){};
    \draw[->, n] (2biF1) -- (2biF2){};
    \draw[->, n] (2biF2) -- (2biF3){};
    \draw[->, n] (2biB2) -- (2biB1){};
    \draw[->, n] (2biB3) -- (2biB2){};

    \draw[->, bl] (s1.north) to [out=70,in=110] (biF1.west){};
    \draw[->, bl] (s2.north) to [out=70,in=110] (biF2.west){};
    \draw[->, n] (s1) -- (biB1){};
    \draw[->, n] (s2) -- (biB2){};

    \draw[->, bl] (s21.north) to [out=70,in=110] (2biF1.west){};
    \draw[->, bl] (s22.north) to [out=70,in=110] (2biF2.west){};
    \draw[->, bl] (s23.north) to [out=70,in=110] (2biF3.west){};
    \draw[->, n] (s21) -- (2biB1){};
    \draw[->, n] (s22) -- (2biB2){};
    \draw[->, n] (s23) -- (2biB3){};

    \node[input] (h1) at (-0.5,3){$h_1$};
    \node[input] (h2) at (1.5,3){$h_2$};
    \node[input] (h21) at (4,3){$h_1$};
    \node[input] (h22) at (6,3){$h_2$};
    \node[input] (h23) at (8,3){$h_3$};

    \node[decoder_block] (decoder) at (3,8.7){};
    \node[rnn] (z1) at (1,8.7) {$z_1$};
    \node[rnn] (z2) at (3,8.7) {$z_2$};
    \node[rnn] (z3) at (5,8.7) {$z_3$};

    \draw[->, n] (z1) -- (z2){};
    \draw[->, n] (z2) -- (z3){};

    \node[block] (u1) at (1,10.3){$u_{1}$};
    \node[block] (u2) at (3,10.3){$u_{2}$};
    \node[block] (u3) at (5,10.3){$u_{3}$};

    \draw[->, n] (z1) -- (u1){};
    \draw[->, n] (z2) -- (u2){};
    \draw[->, n] (z3) -- (u3){};

    \node[input] (y1) at (1,11.5){$y_1$};
    \node[input] (y2) at (3,11.5){$y_2$};
    \node[input] (y3) at (5,11.5){$y_3$};

    \draw[->, n] (u1) -- (z2){};
    \draw[->, n] (u2) -- (z3){};
    \draw[->, n] (u1) -- (y1){};
    \draw[->, n] (u2) -- (y2){};
    \draw[->, n] (u3) -- (y3){};

    \node[attention] (1att) at (-2,6.5){\textsc{att}};
    \node[attention] (2att) at (9,6.5){\textsc{att}};

    \node[cross] (cross1) at (1,4.5){};
    \node[cross] (cross2) at (5,4.5){};
    \node[block] (C1) at (1,5.5){c$_{i}^{(1)}$};
    \node[block] (C2) at (5,5.5){c$_{i}^{(2)}$};
    \node[block] (C3) at (3,6.5){c$_{i}$};

    \node [below=5mm of C3] {Combination};

    \path[->, attedge] (h1) -- node [midway,right] (att1) {} (cross1){};
    \path[->, attedge] (h2) -- node [midway,right] (att2) {} (cross1){};
    \path[->, attedge] (h21) -- node [midway,right] (att21) {} (cross2){};
    \path[->, attedge] (h22) -- node [midway,right] (att22) {} (cross2){};
    \path[->, attedge] (h23) -- node [midway,right] (att23) {} (cross2){};

    \draw[->, alpha] (1att.east) to [out=0,in=180] (att1) {};
    \draw[->, alpha] (1att.east) to [out=0,in=180] (att2) {};
    \draw[->, alpha] (2att.west) to [out=180,in=10] (att21) {};
    \draw[->, alpha] (2att.west) to [out=180,in=10] (att22) {};
    \draw[->, alpha] (2att.west) to [out=180,in=10] (att23) {};

    \draw[->, n] (cross1) -- (C1){};
    \draw[->, n] (cross2) -- (C2){};
    \draw[->, n] (C1) -- (C3){};
    \draw[->, n] (C2) -- (C3){};

    \draw[->, n] (C3.north) to [out=165,in=-95] (z1.south){};
    \draw[->, n] (C3.north) to [out=90,in=-90] (z2.south){};
    \draw[->, n] (C3.north) to [out=15,in=-95] (z3.south){};
    \draw[->, attinput] (z2.south) to [out=205,in=35] (1att.north){};
    \draw[->, attinput] (z2.south) to [out=325,in=135] (2att.north){};
    \draw[->, attinput] (h1.north) to [out=160,in=-120] (1att.south){};
    \draw[->, attinput] (h2.north) to [out=-180,in=-90] (1att.south){};
    \draw[->, attinput] (h21.north) to [out=0,in=-85] (2att.south){};
    \draw[->, attinput] (h22.north) to [out=0,in=-85] (2att.south){};
    \draw[->, attinput] (h23.north) to [out=-310,in=-85] (2att.south){};

\end{tikzpicture}}

\caption{Multi-source S2S with attention. The three combination methods tested are \textsc{concat},
  \textsc{hier} and \textsc{gate}.}
\label{fig:ms}
\end{subfigure}
\caption{The baseline model and the two contextual strategies tested (single and multi-encoder).}
\end{figure*}

In order to correctly translate the type of phenomena mentioned in Sec.~\ref{sec:context},
translation models need to look beyond the sentence. Much of the previous work, mainly in
statistical machine translation (SMT), focused on post-edition, particularly for anaphoric pronoun
translation \cite{guillou_findings_2016,loaiciga_findings_2017}. However, coreference resolution is
not yet sufficient for high quality post- or pre-edition \cite{bawden_cross-lingual_2016}, and for
other discourse phenomena such as lexical cohesion and lexical disambiguation, detecting the
disambiguating context is far from trivial.

Recent work in NMT has explored multi-input models, which integrate the previous sentence as an
auxiliary input. A simple strategy of concatenating the previous sentence to the current sentence
and using a basic NMT architecture was explored by \citet{tiedemann_neural_2017}, but with mixed
results. A variety of multi-encoder strategies have also been tested, including using a
representation of the previous sentence to initialise the main encoder and/or decoder
\cite{wang_exploiting_2017} and using multiple attention mechanisms, with different strategies to
combine the resulting context vectors, such as concatenation \cite{zoph_multi-source_2016},
hierarchical attention \cite{libovicky_attention_2017} and gating \cite{jean_does_2017}.

Although some of the models were evaluated in a contextual setting, for example on the cross-lingal
pronoun prediction task at DiscoMT17 \cite{jean_neural_2017}, certain strategies only appear to give
gains in a low-resource setting \cite{jean_does_2017}, and, more importantly, there has yet to be an
in-depth study into which strategies work best specifically for context-dependent discursive
phenomena.  Here we provide such a study, using the targeted test sets described in
Sec.~\ref{sec:evaluation} to isolate and evaluate the different contextual models' capacity to
exploit extra-sentential context. We test several contextual variants, using both a single encoder
(Sec.~\ref{sec:single}) and multiple encoders (Sec.~\ref{sec:multi}).

\paragraph{NMT notation}
All models presented are based on the widely used encoder-decoder NMT framework with attention
\cite{bahdanau_neural_2015}.  At each decoder step $i$, the context (or summary) vector $c_i$ of the
input sequence is a weighted average of the recurrent encoder states at each input position
depending on the attention weights. We refer to the recurrent state of the decoder as $z_i$. When
multiple inputs are concerned, inputs are noted $x_j^{(k)}$, where $k$ is the input number and $j$
the input position. Likewise, when multiple encoders are used, $c_i^{(k)}$ refers to the
$k$\textsuperscript{th} context vector where $k$ is the encoder number. In the following section,
all $W$s, $U$s and $b$s are learned parameters.

\subsection{Single-encoder models}\label{sec:single}

We train three single-source models: a baseline model and two contextual models. The baseline model
translates sentences independently of each other (Fig.~\ref{fig:baseline}). The two contextual
models, described in \cite{tiedemann_neural_2017}, are designed to incorporate the preceding
sentence by prepending it to the current one, separated by a \textsc{\textless concat\textgreater}
token (Fig.~\ref{fig:concat}).  The first method, which we refer to as \textsc{2-to-2}, is trained
on concatenated source and target sentences, such that the previous and current sentence are
translated together. The translation of the current sentence is obtained by extracting the tokens
following the translated concatenation token and discarding preceding tokens.\footnote{Although the
  non-translation of the concatenation symbol is possible, in practice this was rare ($\textless$0.02\%). If this
  occurs, the whole translation is kept.}  The second method, \textsc{2-to-1}, follows the same
principle, except that only source (and not target) sentences undergo concatenation; the model
directly produces the translation of the current sentence. The comparison of these two methods
allows us to assess the impact of the decoder in producing contextual translations.

\subsection{Multi-encoder models}\label{sec:multi}

Inspired by work on multi-modal translation
\cite{caglayan_multimodal_2016,huang_attention-based_2016}, multi-encoder translation models have
recently been used to incorporate extra-sentential linguistic context in purely textual NMT
\cite{zoph_multi-source_2016,libovicky_attention_2017,wang_exploiting_2017}. Unlike multi-modal
translation, which typically uses two complementary representations of the main input, for example a
textual description and an image, linguistically contextual NMT has focused on exploiting the
previous linguistic context as auxiliary input alongside the current sentence to be
translated. Within this framework, we encode the previous sentence using a separate encoder (with
separate parameters) to produce a context vector of the auxiliary input in a parallel fashion to the
current source sentence. The two resulting context vectors $c^{(1)}_i$ and $c^{(2)}_i$ are then
combined to form a single context vector $c_i$ to be used for decoding (see Fig.~\ref{fig:ms}).  We
study three combination strategies here: concatenation, an attention gate and hierarchical
attention.  We also tested using the auxiliary context to initialise the decoder, similar to
\citet{wang_exploiting_2017}, which was ineffective in our experiments and which we therefore do not
report in this paper.

\paragraph{Attention concatenation}
The two context vectors $c^{(1)}_i$ and $c^{(2)}_i$ are concatenated and the resulting vector
undergoes a linear transformation in order to return it to its original dimension to produce $c_i$ (similar to work
by \newcite{zoph_multi-source_2016}).  
\begin{align}
c_i = W_c [c_i^{(1)};c_i^{(2)}] + b_c
\end{align}

\paragraph{Attention gate}
A gate $r_i$ is learnt between the two vectors in order to give differing importance to the elements of
each context vector, similar to the strategy of \newcite{wang_exploiting_2017}. 
\begin{align}
r_i &= \tanh\left(W_r c_i^{(1)} + W_{s} c_i^{(2)}\right) + b_r \\
c_i &= r_i \odot \left(W_t c_i^{(1)}\right) + \left(1-r_i\right) \odot \left(W_u c_i^{(2)}\right)
\end{align}

\paragraph{Hierarchical attention}
An additional (hierarchical) attention mechanism \cite{libovicky_attention_2017} is introduced to
assign a weight to each encoder's context vector (designed for an arbitrary number of encoders).
\begin{align}
e_i^{(k)} &= v_b^\top \tanh\left(W_b z_{(i-1)} + U_b^{(k)}c_i^{(k)}\right) + b_e \\
\beta_{i}^{(k)} &= \dfrac{\exp\left(e_{i}^{(k)}\right)}{\sum_{k'=1}^K \exp\left(e_{i}^{(k')}\right)} \\
c_i &= \textstyle\sum\nolimits_{k=1}^{K} \beta_i^{(k)} U_c^{(k)} c_i^{(k)}
\end{align}


\subsection{Novel strategy of hierarchical attention and context decoding}
We also test a novel strategy of combining multiple encoders and decoding of both the
previous and current sentence. We use separate, multiple encoders to encode the previous and current
sentence and combine the context vectors using hierarchical attention.  We train
the model to produce the concatenation of the previous and current target sentences, of which the
second part is kept, as in the contextual single encoder models.

\begin{table*}
\centering
\small
\scalebox{0.99}{
    \begin{tabular}{@{}llrrrrrrrr}
      \toprule
      & \multicolumn{4}{c}{\textbf{System Description}} &&\multicolumn{4}{c}{\textbf{\textsc{BLEU} $\uparrow$}} \\
      & \multicolumn{1}{c}{Aux.} & \multicolumn{1}{c}{\#In} &\multicolumn{1}{c}{\#Out} & \#Enc.  & \hphantom{o} & Comedy & Crime & Fantasy & Horror  \\
      
      \addlinespace[2mm]
      \cmidrule(lr){2-5} \cmidrule(lr){7-10}

      \multicolumn{7}{l}{\textit{Single-encoder, non-contexual model}} \\
      \hspace{3mm} \textsc{Baseline} & \ding{55} & 1 & 1 & 1 && 19.52 & 22.07 & 26.30 & 33.05 \\
      \addlinespace[2mm]
      \multicolumn{7}{l}{\textit{Single-encoder with concatenated input}} \\

      \hspace{3mm} \textsc{2-to-2} & src & 2 & 2 & 1 && \ssbest{20.09} & \cellcolor{Green!50}{\textbf{22.93}} & 26.60 & 33.59 \\
      \hspace{3mm} \textsc{2-to-1} & src & 2 & 1 & 1 && 19.51 & 21.81 & 26.78 & \sbest{34.37} \\
      \addlinespace[2mm]


      \multicolumn{7}{l}{\textit{Multi-encoder models (+previous target sentence)}} \\
      \hspace{3mm} \textsc{t-concat} & trg & 2 & 1 & 2 && 18.33 & 20.90 & 24.36 & 32.90 \\
      \hspace{3mm} \textsc{t-hier} & trg & 2 & 1 & 2 && 17.89 & 20.77 & 25.42 & 31.93 \\
      \hspace{3mm} \textsc{t-gate} & trg & 2 & 1 & 2 && 18.25 & 20.76 & 25.55 & 32.64 \\
      \addlinespace[2mm]

      \multicolumn{7}{l}{\textit{Multi-encoder models (+previous source sentence)}} \\
      \hspace{3mm} \textsc{s-concat} & src & 2 & 1 & 2 && 19.35 & 22.41 & 26.50 & 33.67 \\
      \hspace{3mm} \textsc{s-hier} & src & 2 & 1 & 2 && \sbest{20.22} & 21.90 & 26.81 & \ssbest{34.04} \\
      \hspace{3mm} \textsc{s-gate} & src & 2 & 1 & 2 && 19.89 & \ssbest{22.80} & \ssbest{26.87} & 33.81 \\
      \hspace{3mm} \textsc{s-t-hier} & src, trg & 3 & 1 & 3 && 19.53 & 22.53 & \ssbest{26.87} & 33.24  \\
      \addlinespace[2mm]

      \multicolumn{7}{l}{\textit{Multi-encoder with concatenated output}} \\
      \hspace{3mm} \textsc{s-hier-to-2} & src & 2 & 2 & 2  && \best{20.85} & \sbest{22.81} & \sbest{27.17} & \best{34.62}  \\
      \hspace{3mm} \textsc{s-t-hier-to-2} & src, trg & 3 & 2 & 3 && 18.80 & 21.18 & \best{27.68} & 33.33  \\
      \addlinespace[2mm]
      \bottomrule
    \end{tabular}}
    \caption{Results (de-tokenised, cased \textsc{BLEU}) of the ensembled models on four different test sets, each containing three films
      from each film genre. The best, second- and third-best results are highlighted by decreasingly dark shades of green.}
    \label{tab:bleu}
\end{table*} 

\section{Experiments}\label{sec:experiments}

Each of the multi-encoder strategies is tested using the previous source and target sentences as an
additional input (prefixed as \textsc{s-} and \textsc{t-} respectively) in order to test which is
the most useful disambiguating context. Two additional models tested are triple-encoder models,
which use both the previous source and target (prefixed as \textsc{s-t-}). 


\subsection{Data}

Models are trained and tested on fan-produced parallel subtitles from
OpenSubtitles2016\footnote{\url{http://www.opensubtitles.org}}
\cite{lison_opensubtitles2016:_2016}. The data is first corrected using heuristics (e.g. minor
corrections of OCR and encoding errors). It is then tokenised, further cleaned (keeping subtitles
$\leq$80 tokens) and truecased using the Moses toolkit \cite{koehn_moses:_2007} and finally split
into subword units using BPE \cite{sennrich_neural_2016}.\footnote{90,000 merge operations with a
  minimum theshold of 50.} We run all experiments in a high-resource setting, with a training set of
$\approx$29M parallel sentences, with vocabulary sizes of $\approx$55k for English and $\approx$60k
for French.

\subsection{Experimental setup}

All models are sequence-to-sequence models with attention \cite{bahdanau_neural_2015}, implemented
in Nematus \cite{sennrich_nematus:_2017}. Training is performed using the Adam optimiser with a
learning rate of 0.0001 until convergence. We use embedding layers of dimension 512 and hidden
layers of dimension 1024. For training, the maximum sentence length is 50.\footnote{76 when source
  sentences are concatenated to the previous sentence in order to keep the same percentage of
  training sentences as for other models.} We use batch sizes of 80, tied decoder embeddings and
layer normalisation. The hyper-parameters are the same for all models and are the same as those used
for the University of Edinburgh submissions to the news translation shared task at WMT16 and WMT17.  Final models are
ensembled using the last three checkpointed models.

Models that use the previous target sentence are trained using the previous reference translation.
During translation, baseline translations are used. For the targeted evaluation, the problem
does not apply since the translations that are being scored are given.

\section{Results and Analysis}

Overall translation quality is evaluated using the traditional automatic metric \textsc{BLEU}
\cite{papineni_bleu:_2002} (Tab.~\ref{tab:bleu}) to ensure that the models do not degrade overall
performance. We test the models' ability to handle discursive phenomena using the test
sets described in Sec.~\ref{sec:evaluation} (Tab.~\ref{tab:artificial}). The models are
described in the first half of Table~\ref{tab:bleu}: \textit{\#In}~is the number of input sentences, 
the type of auxiliary input of which (previous source or target) is indicated by \textit{Aux.},
\textit{\#Out}~is the number of sentences translated, and \textit{\#Enc}~is the number of encoders
used to encode the input sentences. When there is a single encoder and more than one input, the
input sentences are concatenated to form a single input to the encoder.

\subsection{Overall performance}

Results using the automatic metric \textsc{BLEU} are given in Tab.~\ref{tab:bleu}.  The models are
tested on four different genres of film: comedy, crime, fantasy
and horror.\footnote{Each of the test sets contains three films from that genre, with varying sizes
  and difficulty. The number of sentences in each test set is as follows: comedy: 4,490, crime:
  4,227, fantasy: 2,790 and horror: 2,158.} Scores vary dramatically depending on the genre and the
best model is not always the same for each of the genres.

Contrary to intuition, using the previous target sentence as an auxiliary input (prefix
\textsc{t-}) degrades the overall performance considerably. Testing at decoding time with the
reference translations did not significantly improve this result, suggesting that it is unlikely to
be a case of overfitting during training. The highest performing model is our novel
\textsc{s-hier-to-2} model with more than +1 over the baseline \textsc{BLEU} on almost all test
sets. There is no clear second best model, since performance depends strongly on the test
set used.

\subsection{Targeted evaluation}

Tab.~\ref{tab:artificial} shows the results on the discourse test sets.

\begin{table*}
\small
\centering
\scalebox{0.97}{
      \begin{tabular}{@{}lrrrrrrrrrrr@{}}
        &\multicolumn{7}{c}{\textbf{Coreference (\%)}} &\multicolumn{4}{c}{\textbf{Coherence/cohesion (\%)}}   \\
        & \multicolumn{1}{c}{\textsc{all}} & \multicolumn{1}{c}{\textsc{m.sg.}} & \multicolumn{1}{c}{\textsc{f.sg.}} 
                                     & \multicolumn{1}{c}{\textsc{m.pl.}} & \multicolumn{1}{c}{\textsc{f.pl}} & \textsc{corr.} & \textsc{semi} & 
                                     \hphantom{oooo} & \multicolumn{1}{c}{\textsc{all}} & \hphantom{oo}\\
        
        \addlinespace[2mm]
        \cmidrule(r){2-2}\cmidrule(r){3-6}\cmidrule(r){7-8}\cmidrule(lr){10-10}

        \multicolumn{9}{l}{} \\ 
        \hspace{3mm}\textsc{baseline} & 50.0 & 80.0 & 20.0 & 80.0 & 20.0 & 53.0 & 47.0 
                                                                                                                               && 50.0  \\

        \multicolumn{9}{l}{} \\ 
        \hspace{3mm}\textsc{2-to-2} & \sbest{63.5} & \ssbest{92.0} & \best{50.0} & 84.0 & \sbest{28.0} & \sbest{68.0} & \sbest{59.0}
                                                                                                                               && 52.0 \\
        \hspace{3mm}\textsc{2-to-1} & 52.0 & 72.0 & 28.0 & 84.0 & \ssbest{24.0} & 54.0 & 50.0 
                                                                                                                               && \ssbest{53.0}  \\
        \addlinespace[2mm]


        \multicolumn{7}{l}{} \\ 
        \hspace{3mm}\textsc{t-concat} & 49.0 & 88.0 & 8.0 & \best{96.0} & 4.0 & 50.0 & 48.0 
                                                                                                              && 51.5 \\
        \hspace{3mm}\textsc{t-hier} & 47.0 & 78.0 & 10.0 & \sbest{90.0} & 10.0 & 47.0 & 47.0 
                                                                                                              && 50.5  \\
        \hspace{3mm}\textsc{t-gate} & 47.0 & 80.0 & 6.0 & 82.0 & 20.0 & 45.0 & 49.0 
                                                                                                              && 49.0 \\

        \addlinespace[2mm]
        \multicolumn{7}{l}{} \\ 
        \hspace{3mm}\textsc{s-concat} & 50.0 & 68.0 & 32.0 & 88.0 & 12.0 & \ssbest{55.0} & 45.0 
                                                                                                              && \sbest{53.5} \\
        \hspace{3mm}\textsc{s-hier} & 50.0 & 64.0 & \ssbest{36.0} & 80.0 & 20.0 & \ssbest{55.0} & 45.0 
                                                                                                              && \ssbest{53.0} \\
        \hspace{3mm}\textsc{s-gate} & 50.0 & 68.0 & 32.0 & 84.0 & 16.0 & \ssbest{55.0} & 45.0 
                                                                                                              && 51.5 \\
        \hspace{3mm}\textsc{s-t-hier} & 49.5 & \sbest{94.0} & 4.0 & 88.0 & 12.0 & 53.0 & 46.0 
                                                                          && \ssbest{53.0} \\

        \addlinespace[2mm]
        \multicolumn{7}{l}{} \\
        \hspace{3mm}\textsc{s-hier-to-2} & \best{72.5} & \best{100.0} & \sbest{40.0} & \sbest{90.0} & \best{36.0} & \best{77.0} & \best{68.0} 
                                                                                                              && \best{57.0} \\
        \hspace{3mm}\textsc{s-t-hier-to-2} & \ssbest{56.5} & 84.0 & \ssbest{36.0} & 86.0 & 20.0 & \ssbest{55.0} & \ssbest{58.0} 
                                                                                                              && 51.5 \\

        \bottomrule
      \end{tabular}}
      \caption{Results on the discourse test sets (\% correct). Results on the
        coreference set are also given for each pronoun class. 
        \textsc{corr.} and \textsc{semi} correspond respectively to the ``correct'' and
        ``semi-correct'' examples.
        The best, second- and third-best results are highlighted by decreasingly dark shades of green.}
      \label{tab:artificial}
\end{table*}

\paragraph{Coreference} 

The multi-encoder models do not perform well on the coreference test set; all multi-encoder models
giving at best random accuracy, as with the baseline. This set is designed to test the model's
capacity to exploit previous target context. It is therefore unsurprising that multi-encoder models
using just the previous source sentence perform poorly. It is possible that certain pronouns could
be correctly predicted from the source antecedents, if the antecedent only has one possible
translation. However, this non-robust way of translating pronouns is not tested by
the test set. More surprisingly, the multi-encoder models using the previous target sentence also
perform poorly on the test set. An explanation could be that the target sentence is not being encoded
sufficiently well in this framework, resulting in poor learning. This hypothesis is supported by
the low overall translation performance shown in Tab.~\ref{tab:bleu}.

Two models perform well on the test set: \textsc{2-to-2} and our \textsc{s-hier-to-2}. The high scores,
particularly on the less common feminine pronouns, which can only be achieved through using
contextual linguistic information, show that these models are capable of using previous linguistic
context to disambiguate pronouns. The progressively high performance of these models can be seen in
Fig.~\ref{fig:training_coref}, which illustrates the training progress of these models. The
\textsc{s-t-hier-to-2} model (which uses the previous target sentence as a third auxiliary input)
performs much worse than \textsc{s-hier-to-2}, showing that the addition of the
previous target sentence is detrimental to performance. Whilst the results for the
``correct'' examples (\textsc{corr.}) are almost always higher than the ``semi-correct'' examples
(\textsc{semi}), for which the antecedent is strangely translated, the \textsc{to-2} models also give
improved results on these examples, showing that the target context is necessarily being exploited
during decoding.

These results show that the translation of the previous
sentence is the most important factor in the efficient use of linguistic context.  Combining the
\textsc{s-hier} model with decoding of the previous target sentence (\textsc{s-hier-to-2}) produces
some of the best results across all pronoun types, and the \textsc{2-to-2} model performs almost always
second best.

\pgfplotsset{width=0.5\textwidth,compat=1.9}
\pgfplotsset{every tick label/.append style={font=\scriptsize}}
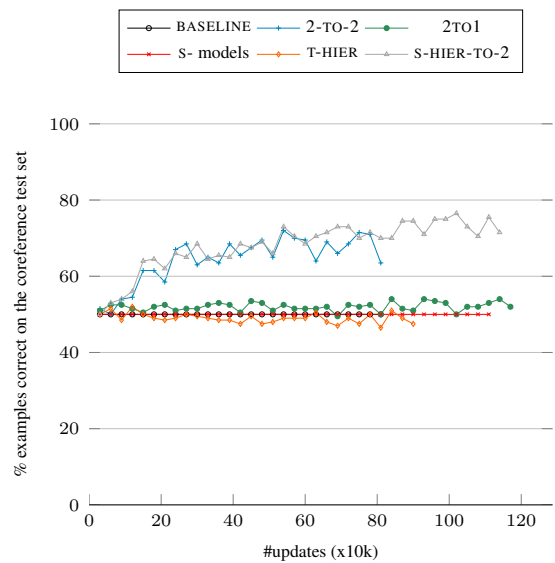
\begin{figure}[ht]
  \scalebox{0.95}{
  \begin{tikzpicture} 
  \begin{axis}[xmin=0, ymin=0, ymax=100,
      axis on top=true, xlabel={\scriptsize\#updates (x10k)}, ylabel={\scriptsize\% examples correct
      on the coreference test set}, 
      legend style={at={(0.5,1.3)}, anchor=north},
      legend columns=3,
      axis x line*=bottom, y axis line style={draw opacity=0}, ymajorgrids]
    
    \addplot[color=Black, mark size=1pt, mark=o]
    coordinates {
      (3,50.0)(6,50.0)(9,50.0)(12,50.0)(15,50.0)(18,50.0)(21,50.0)(24,50.0)(27,50.0)(30,50.0)(33,50.0)(36,50.0)(39,50.0)(42,50.0)(45,50.0)(48,50.0)(51,50.0)(54,50.0)(57,50.0)(60,50.0)(63,50.0)(66,50.0)(69,50.0)(72,50.0)(75,50.0)(78,50.0)(81,50.0)
    }; 
    
    \addplot[color=Blue, mark size=1pt, mark=+] 
    coordinates {
      (3,51.5)(6,50.0)(9,54.0)(12,54.5)(15,61.5)(18,61.5)(21,58.5)(24,67.0)(27,68.5)(30,63.0)(33,65.0)(36,63.5)(39,68.5)(42,65.5)(45,67.5)(48,69.5)(51,65.0)(54,72.0)(57,70.0)(60,69.5)(63,64.0)(66,69.0)(69,66.0)(72,68.5)(75,71.5)(78,71.0)(81,63.5)
    }; 

    \addplot[color=SeaGreen, mark size=1pt, mark=*] 
    coordinates {
      (3,51.0)(6,52.5)(9,52.5)(12,51.5)(15,50.5)(18,52.0)(21,52.5)(24,51.0)(27,51.5)(30,51.5)(33,52.5)(36,53.0)(39,52.5)(42,50.5)(45,53.5)(48,53.0)(51,51.0)(54,52.5)(57,51.5)(60,51.5)(63,51.5)(66,52.0)(69,49.5)(72,52.5)(75,52.0)(78,52.5)(81,50.0)(84,54.0)(87,51.5)(90,51.0)(93,54.0)(96,53.5)(99,53.0)(102,50.0)(105,52.0)(108,52.0)(111,53.0)(114,54.0)(117,52.0)
    }; 

    \addplot[color=red, mark size=1pt, mark=x] 
    coordinates {
      (3,50.0)(6,50.0)(9,50.0)(12,50.0)(15,50.0)(18,50.0)(21,50.0)(24,50.0)(27,50.0)(30,50.0)(33,50.0)(36,50.0)(39,50.0)(42,50.0)(45,50.0)(48,50.0)(51,50.0)(54,50.0)(57,50.0)(60,50.0)(63,50.0)(66,50.0)(69,50.0)(72,50.0)(75,50.0)(78,50.0)(81,50.0)(84,50.0)(87,50.0)(90,50.0)(93,50.0)(96,50.0)(99,50.0)(102,50.0)(105,50.0)(108,50.0)(111,50.0)
    };



    \addplot[color=Orange, mark size=1pt, mark=diamond] 
    coordinates {
      (3,50.0)(6,51.5)(9,48.5)(12,52.0)(15,50.0)(18,49.0)(21,48.5)(24,49.0)(27,50.0)(30,49.5)(33,49.0)(36,48.5)(39,48.5)(42,47.5)(45,49.5)(48,47.5)(51,48.0)(54,49.0)(57,49.0)(60,49.0)(63,50.5)(66,48.0)(69,47.0)(72,49.0)(75,47.5)(78,50.0)(81,46.5)(84,51.0)(87,49.0)(90,47.5)
    };
    
    \addplot[color=Gray, mark size=1pt, mark=triangle] 
    coordinates {
      (3,50.0)(6,53.0)(9,54.0)(12,56.0)(15,64.0)(18,64.5)(21,62.0)(24,66.0)(27,65.0)(30,68.5)(33,64.5)(36,65.5)(39,65.0)(42,68.5)(45,67.5)(48,69.0)(51,66.0)(54,73.0)(57,70.5)(60,68.5)(63,70.5)(66,71.5)(69,73.0)(72,73.0)(75,70.0)(78,71.5)(81,70.0)(84,70.0)(87,74.5)(90,74.5)(93,71.0)(96,75.0)(99,75.0)(102,76.5)(105,73.0)(108,70.5)(111,75.5)(114,71.5)
    };

    \legend{\scriptsize \textsc{baseline}, \scriptsize \textsc{2-to-2}, \scriptsize \textsc{2to1}, 
      \scriptsize \textsc{s-} models, \scriptsize \textsc{t-hier}, 
    \scriptsize \textsc{s-hier-to-2}}
  \end{axis} 
  \end{tikzpicture}}
\caption{Progression of \% correctly ranked examples (from the coreference test set) during training.}
\label{fig:training_coref}
\end{figure}

\paragraph{Coherence and cohesion}
Much less variation in scores can be seen here, suggesting that these examples are more
challenging and that there is room for improvement. Unlike the coreference examples, the
multi-encoder strategies exploiting the previous source sentences perform better than the baseline
(up to 53.5\% for \textsc{s-concat}). Yet again, using the previous target sentence achieves
near random accuracy. \textsc{2-to-2} and \textsc{2-to-1} achieve
similarly low scores (52\% and 53\%), suggesting that if concatenated input is used, decoding the
previous sentence does not add more information. 

However, combining multi-encoding with the decoding of the previous and the current sentences
(\textsc{s-hier-to-2}) greatly improves the handling of the ambiguous translations, improving the
accuracy to 57\%. Extending this same model to also exploit the previous target sentence
(\textsc{s-t-hier-to-2}) degrades this result, giving very similar scores to \textsc{t-hier} and is
therefore not illustrated in FIgure~\ref{fig:training_coref}. This
provides further support for the idea that the target sentence is not encoded efficiently as an
auxiliary input and adds noise to the model, whereas exploiting the target context as a bias in the
recurrent decoder is more effective.

\subsection{How much is the context being used?}

Looking at the attention weights can sometimes offer insights into which input elements are being
attended to at each step.  For coreference resolution, we would expect the decoder to attend to the
pronoun's antecedent. The effect is most expected when the previous target sentence is used, but it
could also apply for the previous source sentence when the antecedent has only one possible
translation. Unlike \citet{tiedemann_neural_2017}, we do not observe increased attention between a
translated pronoun and its source antecedent. Given the discourse test set results, which can only
give high scores when target-side context is used, the contextual information of the type studied in
this paper seems to be best exploited when channelled through the recurrent decoder node rather than
when encoded through the input. This could explain why coreference is not easily
seen via attention weights; the crucial information is encoded on the decoder-side rather than in
the encoder.

\section{Conclusion}

We have presented an evaluation of discourse-level NMT models through the use of two discourse test
sets targeted at coreference and lexical coherence/cohesion. We have shown that multi-encoder
architectures alone have a limited capacity to exploit discourse-level context; poor results are
found for coreference and more promising results for coherence/cohesion, although there is room for
improvement. Our novel combination of contextual strategies greatly outperfoms existing models. This
strategy uses the previous source sentence as an auxiliary input and decodes both the current and
previous sentence. The observation that the decoding strategy is very effective for the
handling of previous context suggests that techniques such as stream decoding, keeping a constant flow of
contextual information in the recurrent node of the decoder, could be very promising for future research.

\section*{Acknowledgments}

Rico Sennrich has received funding from the  Swiss  National  Science Foundation (SNF) in the project CoNTra (grant number 105212\_169888).
This project has also received funding from the European Union’s Horizon 2020 research and innovation
programme under grant agreements 644333 (SUMMA) and 644402 (HimL). 

\FloatBarrier
\bibliography{contextNMT}
\bibliographystyle{acl_natbib}

\end{document}